%% file: all_acl_latex.tex
\pdfoutput=1

\documentclass[11pt]{article}

\usepackage[preprint]{acl}

\usepackage{times}
\usepackage{latexsym}

\usepackage[T1]{fontenc}

\usepackage[utf8]{inputenc}

\usepackage{microtype}

\usepackage{inconsolata}

\usepackage{graphicx}
\usepackage{booktabs}
\usepackage{multirow}
\usepackage[subrefformat=parens]{subcaption}
\usepackage{amsmath}
\usepackage{amsfonts}
\title{Analyzing the Inner Workings of Transformers \\ in Compositional Generalization}

\author{Ryoma Kumon \and Hitomi Yanaka \\
  The University of Tokyo \\
  \texttt{\{kumoryo9, hyanaka\}@is.s.u-tokyo.ac.jp}}

\newcommand{\dobjppiobjpp}[0]{\textsc{pp-iobj}}
\newcommand{\dobjppsubjpp}[0]{\textsc{pp-subj}}

\begin{document}
\maketitle
\begin{abstract}
The compositional generalization abilities of neural models have been sought after for human-like linguistic competence.
The popular method to evaluate such abilities is to assess the models' input--output behavior.
However, that does not reveal the internal mechanisms, and the underlying competence of such models in compositional generalization remains unclear.
To address this problem, we explore the inner workings of a Transformer model by
finding an existing subnetwork that contributes to the generalization performance and by performing causal analyses on how the model utilizes syntactic features.
We find that the model depends on syntactic features to output the correct answer, but that the subnetwork with much better generalization performance than the whole model relies on a non-compositional algorithm in addition to the syntactic features.
We also show that the subnetwork improves its generalization performance relatively slowly during the training compared to the in-distribution one, and the non-compositional solution is acquired in the early stages of the training.
\end{abstract}

\input{sections/1_introduction}

\input{sections/2_background}

\input{sections/3_method}

\input{sections/4_experiment}

\input{sections/5_results}

\input{sections/6_discussion}

\input{sections/7_conclusion}

\input{sections/8_limitation}

\section*{Acknowledgments}
This work was supported by JSPS KAKENHI grant number JP24H00809, JST PRESTO grant number JPMJPR21C8.

\bibliography{anthology, custom}

\appendix

\input{sections/a1_subnetwork_probing}

\input{sections/a3_results_causal_other}

\input{sections/a5_results_subnetwork}

\section{Computational Resources}
We used NVIDIA V100 GPUs for all the experiments.
The total runtime for the training and evaluation was around 300 hours.
\end{document}

%% file: sections/1_introduction.tex
\section{Introduction}
\label{sec:introduction}
Compositional generalization, the ability to understand the meaning of a novel language expression based on the composition of known words and syntactic structures~\citep{partee1984compositionality, fodor1988connectionism}, is a crucial aspect for robustness against unseen language data.
To assess the compositional generalization abilities of neural models, most existing studies have primarily focused on evaluating model outputs in compositional generalization benchmarks~\citep{kim-linzen-2020-cogs, li-etal-2021-compositional, dankers-etal-2022-paradox, li-etal-2023-slog}.
\input{sources/fig1}
However, the model outputs do not necessarily reflect the underlying competence because good performance in the benchmarks does not guarantee that the model implements a solution that generalizes based on compositional rules (i.e., compositional syntax) and vice versa.
In addition, while a growing body of work on model interpretability has investigated the inner workings of Transformer~\citep{vaswani2017attention}-based models~\citep{ferrando2024primerinnerworkingstransformerbased, rai2024practicalreviewmechanisticinterpretability}, compositional generalization has rarely been the focus of such studies.
\citet{yao-koller-2022-structural} and \citet{murty2023characterizing} analyzed the internal mechanisms in compositional generalization but did not focus on the usage of syntactic features, which is the central part of compositional generalization.
Thus, the internal mechanisms of the model in compositional generalization are still unclear, and unveiling them would enhance the understanding of the model's competence.

In this work, we analyze the inner workings of neural models to understand what type of syntactic features the models depend on in tasks requiring compositional generalization.
Our analysis method\footnote{Our code and data are available at \url{https://github.com/ynklab/CG_interp}.} consists of (i) identifying subnetworks within the model that perform well in the generalization and (ii) investigating how syntactic features causally affect the original model and its subnetwork, as shown in Figure~\ref{fig:1}.
The causal analysis involves removing the linguistic concept of interest from the model and comparing the generalization performance before and after the removal.
To rigorously evaluate and analyze the compositional generalization abilities, we focus not on pretrained models but on a Transformer model trained from scratch; this is because pretraining data contain syntactic structures that should be unseen in this experiment, and pretrained models may not need to generalize compositionally~\citep{kim2022uncontrolledlexicalexposureleads}.

We also aim to conduct a detailed analysis by experimenting with various settings.
This study employs two commonly used tasks for evaluating compositional generalization: machine translation and semantic parsing.
We test the models with two patterns of compositional generalization, \dobjppiobjpp{} and \dobjppsubjpp{}, which are similar yet different (see Section~\ref{subsec:cg-pattern} for details).

The findings from our experiments suggest that the model and its subnetwork that contributes to the generalization performance indeed leverage syntactic structures.
However, intriguingly, we also discover that the subnetwork implements other solutions that do not depend on syntactic structures.
From these results, we argue that the solutions that the models employ are partly non-compositional.
Moreover, analysis of the model at different epochs during the training revealed that the model gradually develops a subnetwork with better generalization performance.
The causal analysis of the subnetwork showed that the non-compositional solution was learned during the early phase of the training.
We argue that Transformer models need a better inductive bias to generalize compositionally utilizing syntactic features.

%% file: sources/fig1.tex
\begin{figure}[t]
    \centering
    \includegraphics[width=\linewidth]{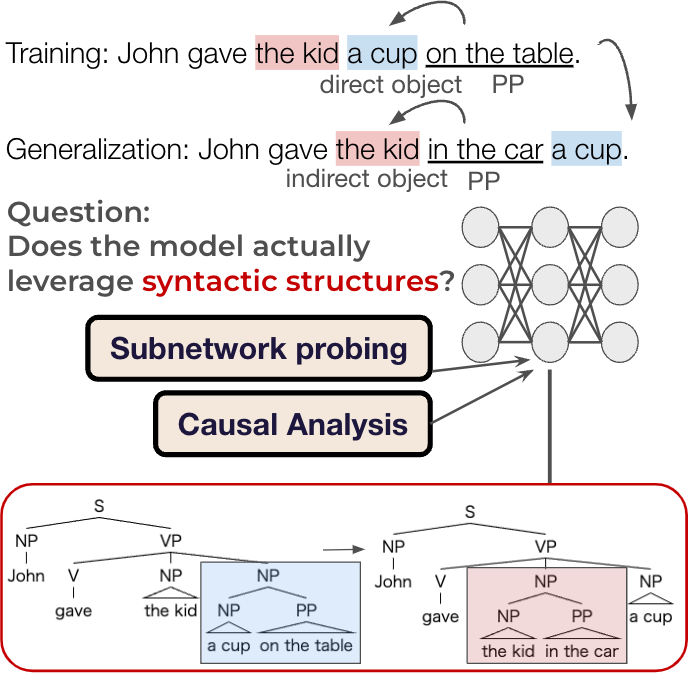}
    \caption{In this study, we investigate what neural models employ in compositional generalization tasks.}
    \label{fig:1}
\end{figure}

%% file: sections/2_background.tex
\section{Background}
\label{sec:background}

\subsection{Compositional Generalization}
\label{subsec:compositional-generalization}
Several studies have explored the compositional generalization abilities of modern neural models by focusing on the model performance in tasks such as semantic parsing~\cite{kim-linzen-2020-cogs, csordas-etal-2021-devil, yao-koller-2022-structural,kim2022uncontrolledlexicalexposureleads, li-etal-2023-slog} and machine translation~\cite{li-etal-2021-compositional, dankers-etal-2022-paradox, kumon-etal-2024-evaluating}.
Such studies indicate that the models lack compositional generalization abilities in general, while others have worked on improving them by modifying the model architectures~\cite{bergan2021edge,ontanon-etal-2022-making}.
\citet{yao-koller-2022-structural} briefly analyzed which part of seq2seq models causes poor performance in compositional generalization by probing the encoder.
They attributed the poor performance to the decoder, stating that the encoder has the linguistic knowledge needed to solve the generalization tasks but the decoder does not use it.
These studies focused mainly on the model outputs and the encoded properties, and there has been a debate on whether behavioral evaluation is sufficient to assess compositionality in neural models~\citep{mccurdy-etal-2024-toward}.
Our research analyzes the inner workings of models with subnetwork search and causal analysis, which should give insights into what features causally impact the model behavior.
We consider that a compositional solution generalizes by leveraging syntactic structures consistently in a bottom-up manner.
In our experiments, we focus on syntactic features that a compositional solution should utilize.
Thus, we regard a solution that does not employ these features as non-compositional.

\subsection{Linguistic Approach to Interpretability}
\label{subsec:interpretability}

One approach to interpreting neural models is to study the causal relationship between the target feature and the model behavior based on interventions.
Interventions alter either a model's inputs or its inner representations so that the target feature is the only change, and they test how that change affects the model's outputs.
Recent work has analyzed the linguistic mechanisms~\citep{tucker-etal-2021-modified, elazar-etal-2021-amnesic, ravfogel-etal-2021-counterfactual, feder-etal-2021-causalm, amini-etal-2023-naturalistic, belrose2023leace, arora-etal-2024-causalgym}.
\citet{belrose2023leace} utilized concept erasure, which removes only the concept of interest from the model and tests causal links by comparing the predictions by the model before and after the removal.
We employ their method for causally analyzing the causal role of syntactic features in compositional generalization because of its compatibility with our compositional generalization tasks.

Another line of work has explored finding subnetworks with specific properties of interest as a method for model analysis.
\citet{cao-etal-2021-low} proposed subnetwork probing, a pruning-based method that searches for a subnetwork that performs a target linguistic task.
As for linguistic generalization, previous studies have found subnetworks that perform syntactic generalization~\citep{bhaskar-etal-2024-heuristic},  hierarchical generalization~\citep{ahuja2024learningsyntaxplantingtrees}, and compositional generalization~\citep{hu2024compositional}.

%% file: sections/3_method.tex
\input{sources/fig2}
\section{Analysis Method}
\label{sec:method}

Figure~\ref{fig:overview} presents our method of analyzing the inner workings of a Transformer model in compositional generalization.
It consists of the following three phases:
training a base model, subnetwork probing, and causal analysis.

\subsection{Training Base Model}
\label{subsec:training}
To test the compositional generalization abilities of a model, we first construct a dataset that contains the training set, in-distribution test set, and out-of-distribution generalization set.
Hereinafter, we refer to the in-distribution test set as the test set and the out-of-distribution generalization set as the generalization set.
The generalization set contains unseen syntactic structures that are combinations of those in the training set and requires models to fill the gaps.
The dataset is constructed based on a rule-based pipeline used in SGET~\citep{kumon-etal-2024-evaluating}, which evaluates the compositional generalization abilities of neural models on English--Japanese translation tasks.
The strict control of sentence generation in SGET utilizing PCFGs (Probabilistic Context-Free Grammars) allows for controlled gaps between the training set and generalization set, which enables the precise evaluation of generalization abilities.

Next, we train a Transformer model from scratch with the training set.
As for training tasks, we adopt machine translation and semantic parsing, which have been commonly used in existing studies of evaluating compositional generalization.
The reason for using two tasks instead of just one is to investigate how the output format of a task impacts the models' inner processes.
The logical forms in the semantic parsing dataset are created mostly based on the rules proposed by \citet{reddy-etal-2017-universal} and postprocessings, following \citet{kim-linzen-2020-cogs}.
We remove redundant tokens while maintaining semantic interpretation, following \citet{wu-etal-2023-recogs}.

\subsection{Subnetwork Probing}
\label{subsec:subnetwork-probing}
Neural models have been shown to develop subnetworks for several types of linguistic generalization~\citep{bhaskar-etal-2024-heuristic, ahuja2024learningsyntaxplantingtrees} and modular solutions for compositionality~\citep{lepori2023break}.
Based on these findings, we hypothesize that the vanilla Transformer develops a subnetwork that generalizes compositionally.
To test this hypothesis, to the trained base model we apply subnetwork probing~\citep{cao-etal-2021-low}, a method for discovering an existing subnetwork that achieves high accuracy on a task.
Subnetwork probing performs pruning-based probing by training a learnable mask.
This method is shown to have low complexity, which means that the mask itself does not learn the task much, and the abilities of the original models are preserved as desired.

In this work, we acquire a subnetwork that performs well in compositional generalization, if any, through subnetwork probing.
We use the generalization set to train masks and prune the models.
The details of subnetwork probing are in Appendix~\ref{sec:subnetwork-probing}.



\input{sources/fig3}
\subsection{Causal Analysis}
\label{subsec:causal}
Next, we analyze the trained models and discovered subnetworks in terms of the extent to which they depend on syntactic structures to generate answers in machine translation and semantic parsing.
One of the methods for analyzing the inner workings is to remove target features from a model and observe the causal effect of the removal.
LEACE~\citep{belrose2023leace} is a method for concept erasure in which only the target concept is removed, with as little impact on the original model as possible.
The method updates inner representations so that no linear classifiers can predict concept labels more accurately than a constant function and other concepts are preserved in the model.
Removing a concept from deep neural networks is achieved by a procedure called concept scrubbing~\citep{belrose2023leace}, which sequentially applies LEACE to every layer of a model from the first to the last.
In particular, after LEACE is applied to a layer and scrubs a concept therein, the scrubbed representations are passed to the next layer, where LEACE is applied again.

We apply concept scrubbing to both the base models and the discovered subnetworks, removing the target syntactic knowledge.
After the concept removal, we evaluate the model predictions in the test set and generalization set of machine translation and semantic parsing.
Comparing the model performances before and after the concept removal reveals the causal effect of the syntactic feature of interest on the predictions.
Concept scrubbing is suitable for our analysis because it does not require creating alternative inputs or interchange interventions, which are difficult to define for controlling syntactic features in the generalization setting.

Concept scrubbing uses a classification task that represents the concept of interest to erase it.
We choose syntactic constituency and syntactic dependency as the concepts for model analysis.
We define multi-label classification tasks to represent each concept based on sequence tagging tasks by \citet{elazar-etal-2021-amnesic}, as shown in Figure~\ref{fig:classification}.
The task for syntactic constituency is tagging the beginning and end of a phrase, and the task for syntactic dependency is labeling dependency relations.
We also test the impact of removing narrower concepts, which differs according to each generalization pattern, such as syntactic constituency regarding only the prepositional phrase (PP) modification of an indirect object noun phrase (NP).
In this case, labels are assigned only to the tokens involved in the concepts.
For example, when considering syntactic constituency regarding the PP modification of an indirect object NP, the beginning and end of the PP and the NP containing the PP and the modified NP (not the modified NP) are labeled as ones.

As for the dataset for this classification task, English sentences are generated using the same rule-based method as the one for the main task datasets.
The labels are tagged based on syntax trees generated as by-products in the sentence generation process.
Constituency boundaries are based on the Penn Treebank~\citep{marcus-etal-1993-building} definitions, and dependency relations are based on the Universal Dependencies~\citep{mcdonald-etal-2013-universal} definitions.
Using these tasks and datasets, we test whether the models depend on syntactic constituency and syntactic dependency in compositional generalization.

%% file: sources/fig2.tex
\begin{figure*}[t]
    \centering
    \includegraphics[width=\linewidth]{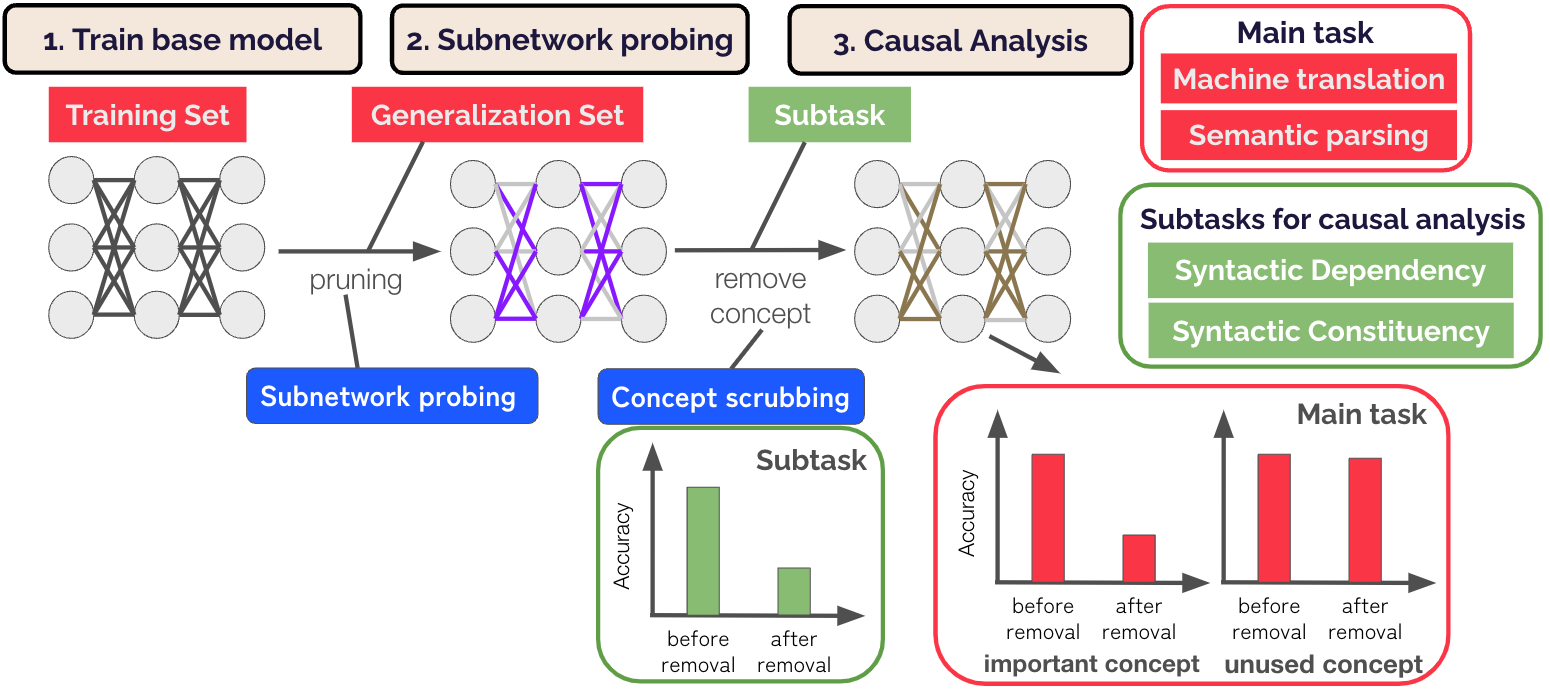}
    \caption{Overview of analysis process, which consists of three phases: training base models, subnetwork probing, and causal analysis.}
    \label{fig:overview}
\end{figure*}

%% file: sources/fig3.tex
\begin{figure}[t]
    \centering
    \includegraphics[width=0.95\linewidth]{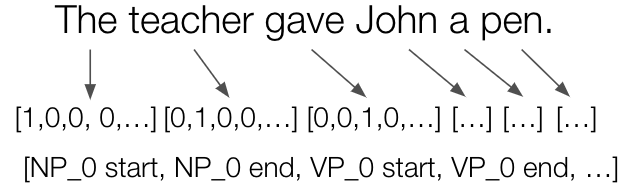}
    \caption{Multi-label classification task used in concept scrubbing for removing syntactic constituency.}
    \label{fig:classification}
\end{figure}

%% file: sections/4_experiment.tex
\section{Experimental settings}
\label{sec:experiment}

\input{sources/tab_generalization_pattern}

\subsection{Compositional Generalization Pattern}
\label{subsec:cg-pattern}

In this work, we focus on two compositional generalization patterns, i.e., PP in indirect object NP (\dobjppiobjpp{}) and PP in subject NP (\dobjppsubjpp{}{}), as shown in Table~\ref{tab:generalization_pattern}.
Syntactically, these two patterns are relatively simple, which allows for easier causal analyses using multi-label classification tasks.

\paragraph{PP in indirect object NP (\dobjppiobjpp{})}
In this pattern, all the NPs modified by PPs in the training set appear in the direct object position.
Then, models trained on the training set are expected to generalize to PPs modifying indirect object NPs in the generalization set.
As \citet{li-etal-2023-slog} pointed out, some sentences have an indirect object NP modified by a PP before a direct object NP, and the dependency between a verb and the direct object NP goes across the PP, which makes the generalization more complex.

For causal analysis in this pattern, we test the impact of three narrower syntactic constituency and dependency concepts, namely, the PP modification of indirect object NPs, direct object NPs, and all NPs, along with overall syntactic constituency and dependency.
\paragraph{PP in subject NP (\dobjppsubjpp{})}
Similarly to \dobjppiobjpp{}, all the NPs modified by PPs in the training set appear in the direct object position.
Models trained on the training set are expected to generalize to PPs modifying subject NPs in the generalization set.
One aspect that makes \dobjppsubjpp{} difficult is that PP modifiers do not appear at the beginning of sentences in the training set.
The models may have to generalize to the novel placement of PP modifiers in addition to the novel grammatical role of modified NPs.
To mitigate this issue, \citet{wu-etal-2023-recogs} added sentences with preposed PP modifiers in the training set, but we avoid that approach for the sake of simpler comparisons between \dobjppiobjpp{} and \dobjppsubjpp{}.

For causal analysis in this pattern, we test the impact of three narrower syntactic constituency and dependency concepts, namely, the PP modification of subject NPs, direct object NPs, and all NPs, along with overall syntactic constituency and dependency.


\subsection{Dataset}
As explained in Section~\ref{sec:method}, we newly construct datasets for each of machine translation and semantic parsing and for classification tasks used in concept scrubbing, using PCFGs with vocabulary of 123 proper nouns, 423 common nouns, 178 verbs, and 43 adjectives.
Each of the machine translation and semantic parsing datasets consists of a training set of 80,000 samples, a test set of 10,000 samples, and a generalization set of 30,000 samples.
We split the generalization set into two parts:
one part is used in training masks in subnetwork probing (Section~\ref{subsec:subnetwork-probing}), and the other is used in evaluating the trained models and subnetworks (Section~\ref{subsec:causal}).
Note that the generalization set is constructed for each generalization pattern, and subnetwork probing is performed for each pattern as well.
The dataset for classification tasks contains 9,000 samples, and all of them are used for concept scrubbing.

\subsection{Training Details}
We train an encoder--decoder Transformer model from scratch on our dataset.
The model has 3 encoder and 3 decoder layers, 4 attention heads.
We set the batch size to 256, the number of epochs to 500, the learning rate to 0.0001, and the weight decay to 0.1.
We do not use early stopping because \citet{csordas-etal-2021-devil} showed that continued training without it improves model performance in compositional generalization.

As for subnetwork probing, we train a pruning mask for the trained model.
We set the batch size to 256, the number of training epochs to 300, and the learning rate to 0.0005.
We do not use early stopping in subnetwork probing.

We run the experiments three times with random seeds and report the average scores as the results.
The final checkpoints of each training run are used for the main results (Section~\ref{sec:results}).

\subsection{Evaluation Metric}
\label{subsec:metric}
Following previous studies of evaluating compositional generalization~\citep{kim-linzen-2020-cogs, kumon-etal-2024-evaluating}, we adopt exact match accuracy as the evaluation metrics for both machine translation and semantic parsing.
The rule-based pipeline for our dataset generation is designed so that a correct output can be determined uniquely if a model follows compositional rules; thus, using exact match accuracy in this experiment is appropriate.

%% file: sources/tab_generalization_pattern.tex
\begin{table*}[t]
    \centering
    \begin{tabular}{p{0.20\linewidth}p{0.37\linewidth}p{0.35\linewidth}}
    \toprule
    Pattern & Training & Generalization \\
    \midrule
    PP in indirect object NP&
     The child gave \textbf{the pen on the table} to Liam.&
     The friend gave \textbf{the girl in the room} a hat.\\
    PP in subject NP &
     The child broke \textbf{a cup on the table}. &
    \textbf{The friend in the room} broke a cup.\\
     \bottomrule
    \end{tabular}
    \caption{Two compositional generalization patterns tested in the experiments.}
    \label{tab:generalization_pattern}
\end{table*}

%% file: sections/5_results.tex
\section{Results}
\label{sec:results}
\input{sources/tab_results_overall}
\input{sources/fig_results_causal_dobjpp2iobjpp}
\subsection{Output Evaluation}
\label{subsec:overall}
Before analyzing the inner workings of the models, we present the model performance in main tasks without concept scrubbing.
Table~\ref{tab:results_overall} shows the results of the base models and subnetworks in machine translation and semantic parsing.

The base models and subnetworks both performed nearly perfectly in the test set of both machine translation and semantic parsing.
The performance of the base model was much worse in both \dobjppiobjpp{} and \dobjppsubjpp{} than that in the test set, which is consistent with the results of previous studies testing the same generalization patterns~\citep{li-etal-2023-slog, kumon-etal-2024-evaluating}.
On the other hand, the subnetwork scored more than 90\% accuracy in \dobjppiobjpp{} in both main tasks while keeping the in-distribution performance.
This suggests that some part of the trained model implements a certain algorithm that solves these compositional generalization tasks.
The subnetwork in \dobjppsubjpp{} also performed much better in the generalization set than did the base model.
It is surprising to see these positive results, considering that previous studies have shown Transformers' poor performance in compositional generalization tasks.

\subsection{Causal Analysis}
\label{subsec:results_causal}
\input{sources/fig_results_overall_epoch}
\subsubsection{Generalization in \dobjppiobjpp{}}
Figures~\ref{fig:results_causal_a}--\ref{fig:results_causal_d} present the results of causal analysis of the base model and subnetwork in \dobjppiobjpp{}.
First, the base model performed much worse when syntactic constituency or dependency was removed, which shows the model's reliance on these syntactic features to correctly solve the main tasks.
However, the base model cannot be considered to have a compositionally generalizing solution because the generalization performance overall was far from perfect, and a compositionally generalizing model should perform nearly perfectly in the generalization set.

Next, we focus on the subnetwork, which achieved better accuracy in the generalization set than did the base model.
Entirely removing syntactic constituency or dependency decreased the accuracy of the subnetwork to almost zero except when removing syntactic dependency in machine translation.
This shows that the subnetwork also depends on the syntactic features in general.

We then discuss the impact of the removal of constituency information on the modification of indirect object NPs.
If the subnetwork implements a compositional solution, then removing the constituency information on the modification of indirect object NPs would decrease the accuracy to zero.
However, the difference in the performance before and after this concept removal is not as large as when the concept of the constituency is removed entirely, although the generalization performance of the subnetwork decreases to some extent.
A similar trend was seen in the removal of syntactic dependency, although the decline in the generalization performance was smaller.
These results suggest that the subnetwork depends somewhat on the constituency and dependency regarding the modification of indirect object NPs.
At the same time, the subnetwork implements a solution that somehow handles \dobjppiobjpp{} in machine translation and semantic parsing yet cannot be considered as a compositionally generalizing one.

Furthermore, regarding the differences in the results between the two tasks, the drop in the generalization performance after the removal of syntactic information was smaller in semantic parsing than in machine translation.
It indicates that the models relied on syntactic features instead of a non-compositional solution more in semantic parsing than in machine translation.

\input{sources/fig_results_causal_epoch}
\subsubsection{Generalization in \dobjppsubjpp{}}
In contrast to \dobjppiobjpp{}, the generalization performance of the subnetwork is far from 100\% accuracy, so the subnetwork is not expected to have a perfect solution that generalizes compositionally in \dobjppsubjpp{}.
However, we still examine on what the subnetwork depends in the main tasks.

Figures~\ref{fig:results_causal_e}--\ref{fig:results_causal_h} show the results of causal analysis of the subnetwork in \dobjppsubjpp{}.
Similar to \dobjppiobjpp{}, when the information of syntactic constituency or dependency was entirely removed, the performance of the subnetwork dropped to almost 0\% except when removing syntactic dependency in machine translation.
The subnetwork depends on this information in \dobjppsubjpp{} as well.

As for the impact of the removal of the constituency regarding the modification of subject NPs, the performance dropped almost as much as with the removal of the entire constituency.
This suggests the subnetworks' heavy reliance on the modification of subject NPs and their ability to properly use compositional rules at least when the output is correct.
On the other hand, when the constituency regarding direct object NPs was removed, the performance improved considerably especially in semantic parsing.
This implies that the subnetwork was overfitted to the modification of direct object NPs, and this prevented the subnetwork from achieving better accuracy in \dobjppsubjpp{}.
Also, the subnetwork seems capable of using the information of modification of NPs regardless of whether NPs are subjects or direct objects, as this removal would not improve the accuracy if the subnetwork learns only that the modification can only come with direct object NPs.

%% file: sources/tab_results_overall.tex
\begin{table}[t]
    \centering
    \small
    \begin{tabular}{p{0.05\linewidth}p{0.22\linewidth}p{0.14\linewidth}p{0.14\linewidth}p{0.15\linewidth}}\toprule
    Task & Model & Test & \dobjppiobjpp{} & \dobjppsubjpp{}\\\midrule
    \multirow{3}{*}{MT} & Base & $99.9_{\pm 0.0}$ & $47.0_{\pm 2.2}$ & $0.0_{\pm 0.0}$  \\
    & \dobjppiobjpp{} Sub. & $99.8_{\pm 0.0}$ & $91.4_{\pm 3.0}$ & ---\\
    & \dobjppsubjpp{} Sub. & $96.5_{\pm 3.5}$ & --- & $57.0_{\pm 18.8}$ \\
    \multirow{3}{*}{SP} & Base & $99.8_{\pm 0.0}$ & $55.3_{\pm 4.1}$ & $0.1_{\pm 0.0}$\\
    & \dobjppiobjpp{} Sub. & $94.3_{\pm 0.8}$ & $91.2_{\pm 5.5}$ & ---\\
    & \dobjppsubjpp{} Sub. & $98.3_{\pm 0.0}$ & --- & $12.4_{\pm 7.4}$\\\bottomrule
    \end{tabular}
    \caption{Average exact match accuracy (\%) of base models and subnetworks in machine translation (MT) and semantic parsing (SP).
    \dobjppiobjpp{} Sub. (resp. \dobjppsubjpp{} Sub.) stands for the subnetwork for \dobjppiobjpp{} (resp. \dobjppsubjpp{}).
    The column labeled \dobjppiobjpp{} (resp. \dobjppsubjpp{}) represents the generalization performance in \dobjppiobjpp{} (resp. \dobjppsubjpp{}).}
    \label{tab:results_overall}
\end{table}

%% file: sources/fig_results_causal_dobjpp2iobjpp.tex
\begin{figure*}[t]
    \centering
    \begin{minipage}[b]{0.245\linewidth}
    \centering
    \includegraphics[width=\linewidth]{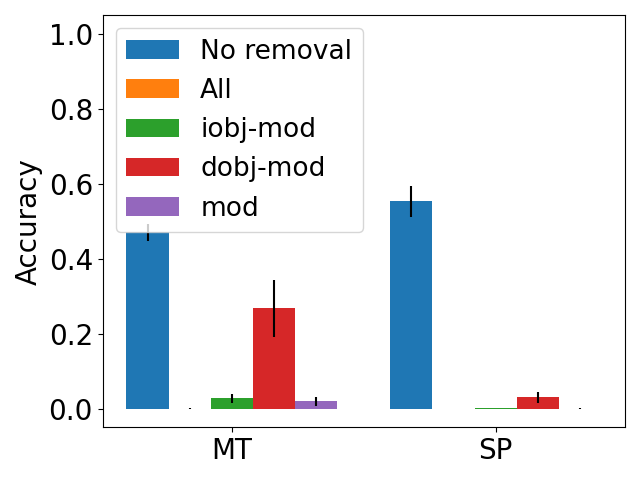}
    \subcaption{Constituency (Base)}
    \label{fig:results_causal_a}
    \end{minipage}
    \begin{minipage}[b]{0.245\linewidth}
    \centering
    \includegraphics[width=\linewidth]{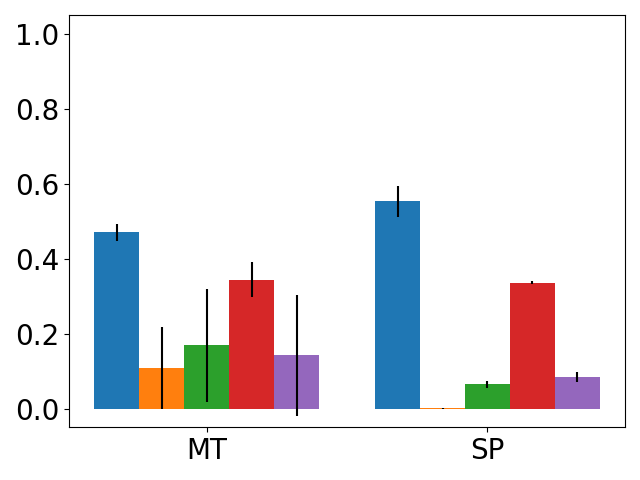}
    \subcaption{Dependency (Base)}
    \label{fig:results_causal_b}
    \end{minipage}
    \begin{minipage}[b]{0.245\linewidth}
    \centering
    \includegraphics[width=\linewidth]{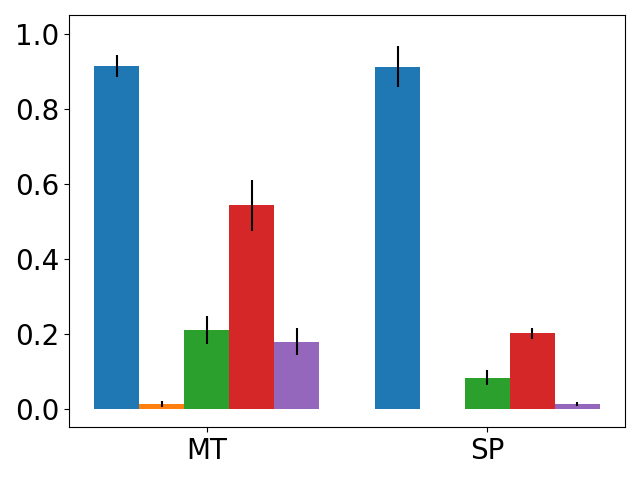}
    \subcaption{Constituency (Sub.)}
    \label{fig:results_causal_c}
    \end{minipage}
    \begin{minipage}[b]{0.245\linewidth}
    \centering
    \includegraphics[width=\linewidth]{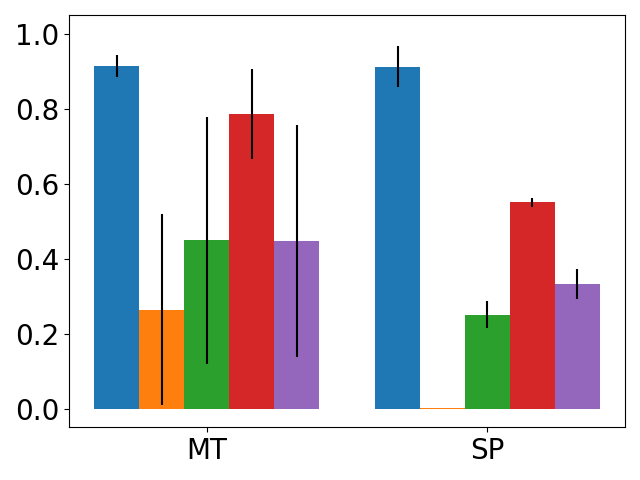}
    \subcaption{Dependency (Sub.)}
    \label{fig:results_causal_d}
    \end{minipage}
    \begin{minipage}[b]{0.245\linewidth}
    \centering
    \includegraphics[width=\linewidth]{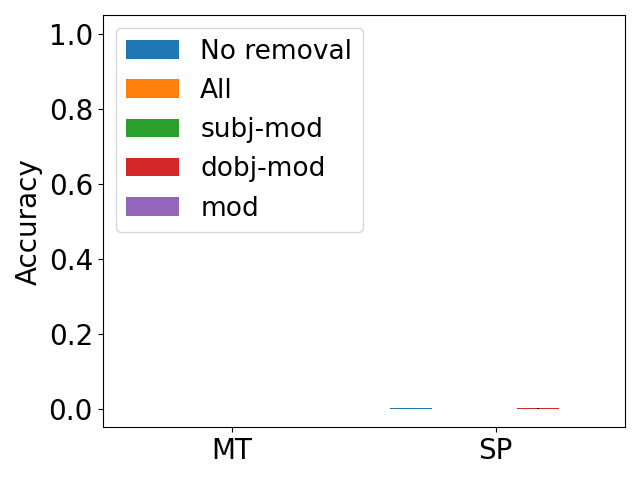}
    \subcaption{Constituency (Base)}
    \label{fig:results_causal_e}
    \end{minipage}
    \begin{minipage}[b]{0.245\linewidth}
    \centering
    \includegraphics[width=\linewidth]{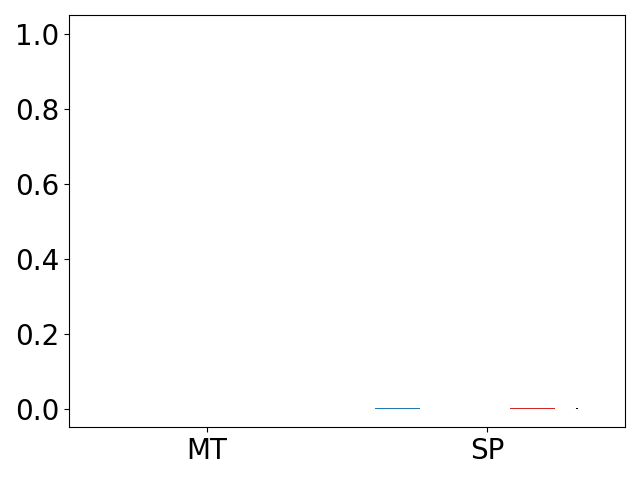}
    \subcaption{Dependency (Base)}
    \label{fig:results_causal_f}
    \end{minipage}
    \begin{minipage}[b]{0.245\linewidth}
    \centering
    \includegraphics[width=\linewidth]{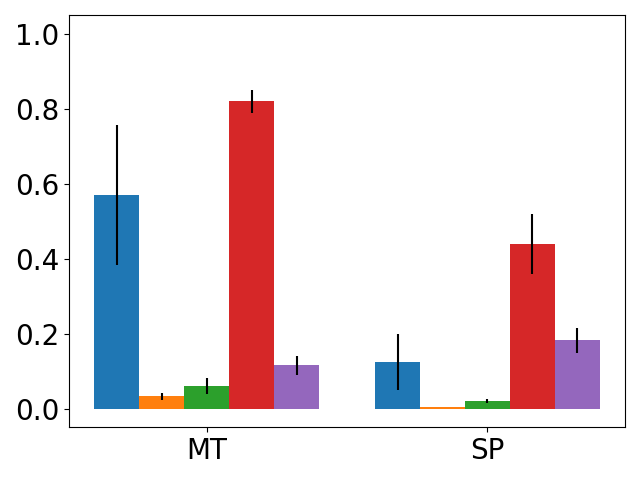}
    \subcaption{Constituency (Sub.)}
    \label{fig:results_causal_g}
    \end{minipage}
    \begin{minipage}[b]{0.245\linewidth}
    \centering
    \includegraphics[width=\linewidth]{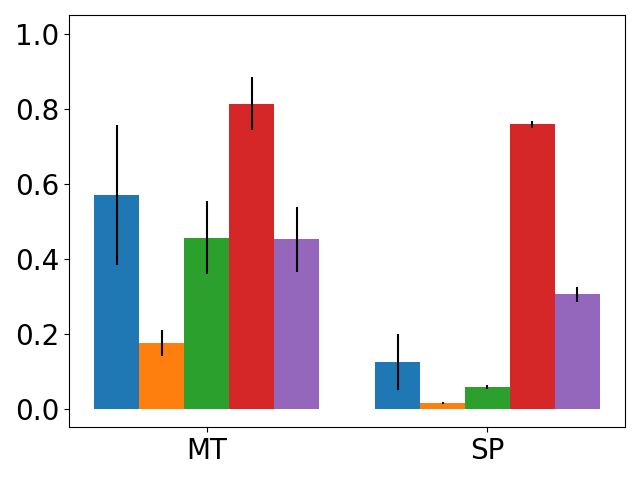}
    \subcaption{Dependency (Sub.)}
    \label{fig:results_causal_h}
    \end{minipage}
    \caption{Results of causal analysis in \dobjppiobjpp{} (\ref{fig:results_causal_a}-\ref{fig:results_causal_d}) and in \dobjppsubjpp{} (\ref{fig:results_causal_e}-\ref{fig:results_causal_h}).
    Each bar shows the performance in the generalization set after the corresponding concept removal.
    ``All" refers to entirely removing the corresponding syntactic feature.
    ``Iobj-mod" (resp. ``dobj-mod", ``subj-mod") refers to removing the corresponding syntactic feature regarding the PP modifications of indirect object (resp. direct object, subject) NPs.
    ``Mod" refers to removing the corresponding syntactic feature regarding the PP modifications.}
    \label{fig:results_causal}
\end{figure*}

%% file: sources/fig_results_overall_epoch.tex
\begin{figure*}[t]
    \centering
    \begin{minipage}[b]{0.35\linewidth}
    \centering
    \includegraphics[width=\linewidth]{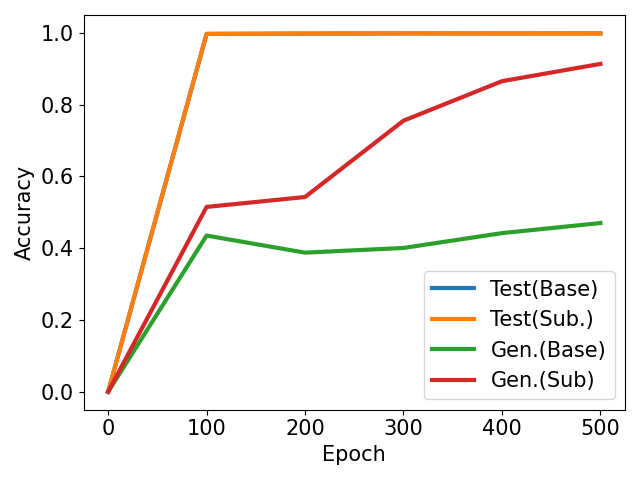}
    \subcaption{Machine translation}
    \end{minipage}
    \begin{minipage}[b]{0.35\linewidth}
    \centering
    \includegraphics[width=\linewidth]{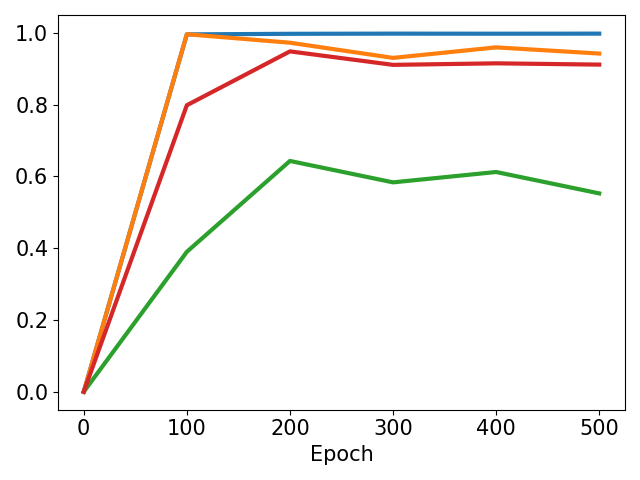}
    \subcaption{Semantic parsing}
    \end{minipage}
    \caption{Shift of average accuracy of models in \dobjppiobjpp{} over training epochs.
    Test and Gen. stand for the accuracy on the test and generalization set, respectively.
    Base and Sub. stand for the accuracy of the base model and subnetwork, respectively.}
    \label{fig:results_overall_epoch}
\end{figure*}

%% file: sources/fig_results_causal_epoch.tex
\begin{figure*}[t]
    \centering
    \begin{minipage}[b]{0.245\linewidth}
    \centering
    \includegraphics[width=\linewidth]{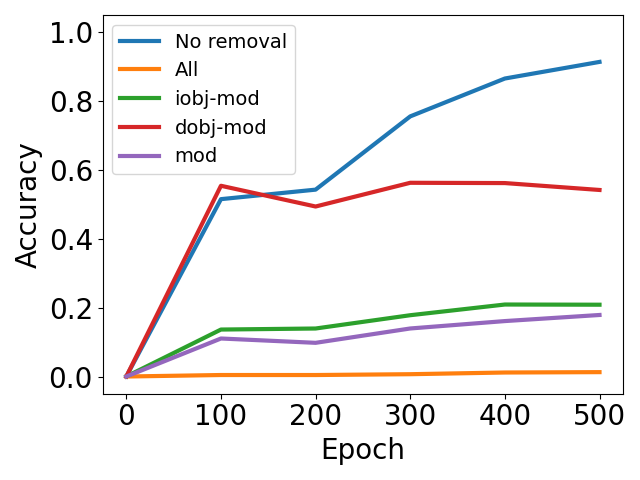}
    \subcaption{Constituency (\dobjppiobjpp{})}
    \end{minipage}
    \begin{minipage}[b]{0.245\linewidth}
    \centering
    \includegraphics[width=\linewidth]{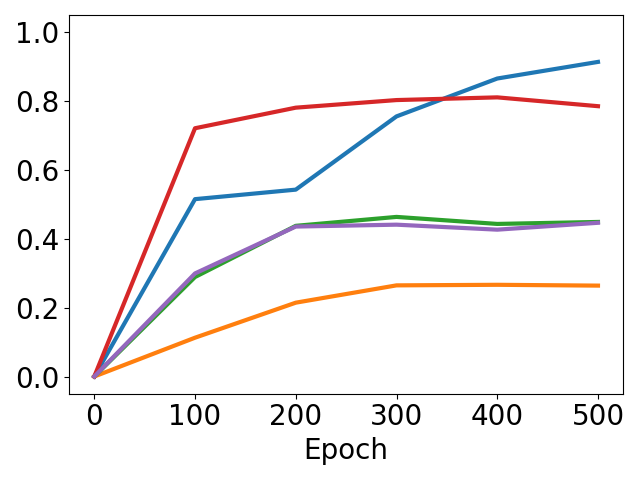}
    \subcaption{Dependency (\dobjppiobjpp{})}
    \end{minipage}
    \begin{minipage}[b]{0.245\linewidth}
    \centering
    \includegraphics[width=\linewidth]{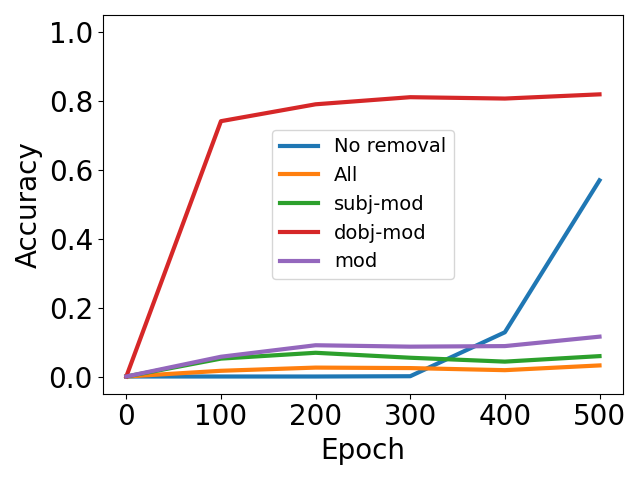}
    \subcaption{Constituency (\dobjppsubjpp{})}
    \end{minipage}
    \begin{minipage}[b]{0.245\linewidth}
    \centering
    \includegraphics[width=\linewidth]{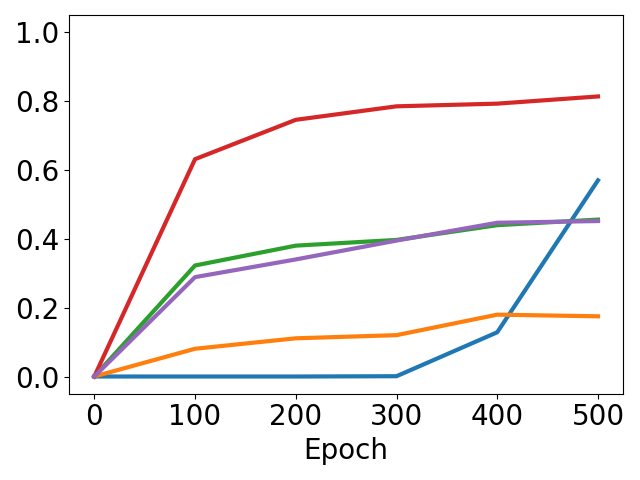}
    \subcaption{Dependency (\dobjppsubjpp{})}
    \end{minipage}
    \caption{Shift of average accuracy on the generalization set of the subnetworks with each concept removed in machine translation over training epochs.}
    \label{fig:results_causal_epoch}
\end{figure*}

%% file: sections/6_discussion.tex

\subsection{Transition During Training}
We present how the model performance evolved throughout training in \dobjppiobjpp{}, shown in Figure~\ref{fig:results_overall_epoch}.
As can be seen, the accuracy in the test set grew rapidly, whereas the accuracy in the generalization set improved slowly.
In addition, comparing the base model and subnetwork in machine translation, the generalization performance of the subnetwork continued to improve through 500 epochs, whereas that of the base model improved only slightly.
Therefore, the model may gradually learn an algorithm that can solve the generalization task through the training process with the machine translation task without changing the behavior of the whole model much.
The difference in the performance transition between the base model and subnetwork is much less noticeable in semantic parsing.
These results suggest that differences in the task settings---such as output formats where structures are represented more explicitly in semantic parsing---influence generalization performance.
A similar tendency was discovered in \dobjppsubjpp{} (see Appendix~\ref{sec:causal_other} for details), although the generalization performance was generally lower.

Next, we investigate how the inner workings of the subnetwork changed as the training went on.
Figure~\ref{fig:results_causal_epoch} shows the shift of the generalization performance in machine translation of the subnetwork with each linguistic feature removed.
In machine translation, the generalization performance of the subnetwork with certain syntactic feature removed was mostly consistent after 200 epochs.
This strongly suggests that the subnetwork learned a non-compositional solution in the early stage of the training and retained it throughout the training.
Combined with the observation that the original subnetwork continued to improve its accuracy beyond 200 epochs, this result also indicates that a compositional solution relying on syntactic features was acquired gradually.

Similarly, in semantic parsing, the subnetwork with a certain linguistic feature removed mostly retained its performance after 200 epochs, regardless of the generalization accuracy of the original subnetwork.
The detailed results of semantic parsing are in Appendix~\ref{sec:causal_other}.

\section{Discussion}
\label{sec:discussion}
\subsection{Reliability of LEACE}
\input{sources/tab_results_word}
Since LEACE is a linear concept erasure method, it may fail to remove concepts encoded non-linearly.
However, the results when all syntactic information was removed (Figure~\ref{fig:results_causal}) indicate that most syntactic information used in machine translation and semantic parsing is encoded in linear subspaces.
Furthermore, we use linear probing to assess whether the concepts are perfectly removed after the application of concept scrubbing.
We probe the representation of the final layer of the encoder after each concept scrubbing, and measure the accuracy in the multi-label classification tasks used in concept scrubbing.
As a result, the probing classifier predicted the correct labels for all the words in 0\% of the test sentences in almost all combination of the removed concepts, generalization patterns, and main tasks.
The only exceptions occurred when either all syntactic dependencies or those related to the PP modification of all NPs were removed, with maximum accuracies of 9.7\% and 1.3\%, respectively.
Thus, the impact of non-linearly encoded features should be negligible, and concept scrubbing effectively removes syntactic features.

We also validate that LEACE does not erase concepts orthogonal to syntactic ones.
We test the model with a word-to-word translation (English to Japanese) of content words; word-to-word translation of content words can be solved without relying on syntax at all.
We probe the representation of the final layer of the encoder by a one-layer linear classifier for word-to-word translation, focusing on evaluating the models trained with machine translation datasets. Accuracy is calculated as the proportion of sentences in which all content words are translated correctly.
As shown in Table~\ref{tab:results_word}, the results suggest that the removal of syntactic features only slightly decreases performance in word-to-word translation, confirming that concepts orthogonal to syntactic ones are mostly preserved in LEACE.

\subsection{Impact of Adding Hints in Training}
\input{sources/fig_results_hint}
Finally, we investigate how a Transformer model performs under a setting where compositional generalization is easier.
We augment the training set with sentences containing syntactic structures that provide clues for generalization.
In particular, we focus on \dobjppiobjpp{} and augment the training set with sentences that have a relative clause (RC) modifying an indirect object NP and with ones that have an RC modifying a direct object NP.
It should be easier for the model trained with the data involving PPs modifying direct object NPs to generalize to a PP modifying an indirect object NP based on the newly provided hints.

Figure~\ref{fig:results_hint} presents the results of causal analysis.
Compared with the results without any hint (Figures~\ref{fig:results_causal_a} and \ref{fig:results_causal_b}), the generalization performance without any concept removal improved by about 40\% in both main tasks.
Moreover, the performance after the removal of syntactic features regarding the PP modifications of indirect object NPs improved only slightly.
This suggests that the algorithm that was implemented in the base model and contributed to the gain in the generalization performance relied on those specific syntactic features.
Thus, the model might implement a more robust compositional solution by utilizing the provided hints.

%% file: sources/tab_results_word.tex
\begin{table}[t]
    \centering
    \small
    \begin{tabular}{p{0.23\linewidth}p{0.17\linewidth}p{0.19\linewidth}p{0.19\linewidth}}\toprule
    Model & Original & Constituency removed & Dependency removed \\\midrule
    Base & $90.1_{\pm 2.8}$ & $88.8_{\pm 4.6}$ & $90.0_{\pm 3.6}$\\
    \dobjppiobjpp{} Sub. & $87.9_{\pm 4.8}$ & $83.8_{\pm 6.3}$ & $89.3_{\pm 3.0}$\\ 
    \dobjppsubjpp{} Sub. & $88.0_{\pm 2.1}$ & $81.2_{\pm5.4}$ & $88.2_{\pm 2.3}$ \\
    \bottomrule
    \end{tabular}
    \caption{Average accuracy in word-to-word translation of content words.}
    \label{tab:results_word}
\end{table}

%% file: sources/fig_results_hint.tex
\begin{figure}[t]
    \centering
    \begin{minipage}[b]{0.49\linewidth}
    \centering
    \includegraphics[width=\linewidth]{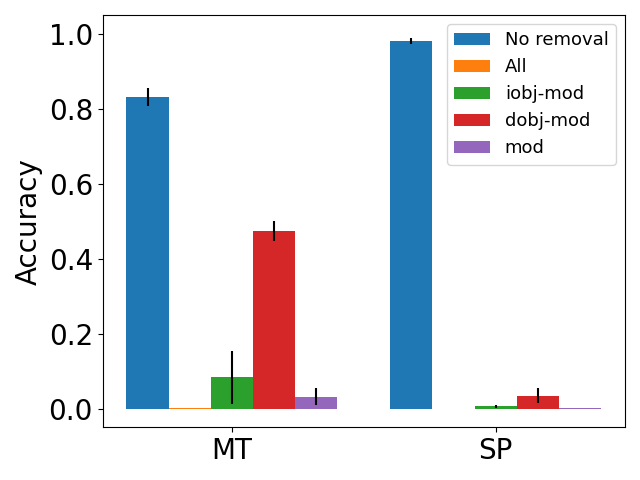}
    \subcaption{Constituency (Base)}
    \end{minipage}
    \begin{minipage}[b]{0.49\linewidth}
    \centering
    \includegraphics[width=\linewidth]{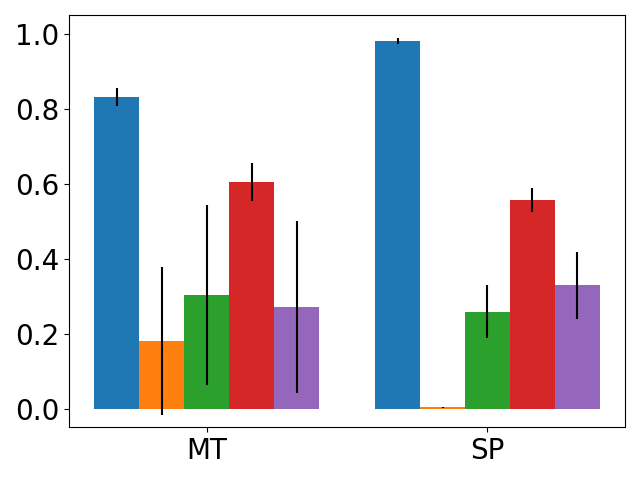}
    \subcaption{Dependency (Base)}
    \end{minipage}
    \caption{Results of causal analysis in \dobjppiobjpp{} when trained with a hint.}
    \label{fig:results_hint}
\end{figure}

%% file: sections/7_conclusion.tex
\section{Conclusion}
\label{sec:conclusion}
In this work, we investigated the inner mechanisms of a Transformer model in compositional generalization tasks.
The experimental results showed that the model utilizes syntactic features to some extent in the generalization but that its subnetwork with better generalization accuracy depends on non-syntactic features as well.
This indicates that the model develops a non-compositional solution internally and fails to generalize compositionally even when the generalization performance is decent.
This paper serves as a foundation for analyzing the underlying mechanisms in compositional generalization from the linguistic perspective.
Future work might consider other generalization patterns or other linguistic features to obtain a more profound insight into the linguistic competence of neural models.

%% file: sections/8_limitation.tex
\section*{Limitations}
The compositional generalization tasks used in this work are based on synthetically generated datasets and so might not represent sufficiently the variety in natural language expressions.
However, these controlled settings are required for precise evaluation requires because all the lexical items and syntactic structures must be split properly into the training and generalization sets.
Therefore, using a natural corpus for this experiment would have required much effort, and we leave that for future work.

Another limitation is that the results of this experiment do not necessarily transfer to larger models because we tested relatively small models trained on a small synthetic dataset following previous studies.
It would be worth exploring how the trends discovered here change as the model size increases.

\section*{Ethical Considerations}
All of our datasets were constructed for the sole purpose of the model analysis from the linguistic perspective.
They contain no potentially harmful or offensive content.

%% file: sections/a1_subnetwork_probing.tex
\section{Subnetwork Probing}
\label{sec:subnetwork-probing}
Subnetwork probing~\citep{cao-etal-2021-low} trains a mask to find a subnetwork of interest.
Let $\phi \in \mathbb{R}^d$ be the weights of a model and $Z_{i}\in [0,1]$ be the mask for the weight $\phi_i$.
$Z_i$ follows the hard concrete function parameterized with temperature $\beta_i$ and a random variable $\theta_i$, that is,
\begin{align*}
U_i &\sim \mathrm{Unif}[0, 1],\\
S_i &= \sigma\left(\frac{1}{\beta}\left(\log\frac{U_i}{1-U_i}+\theta_i\right)\right),\\
Z_i &= \min (1, \max(0, S_i(\zeta-\gamma)+\gamma)),
\end{align*}
and $\zeta=1.1$ and $\gamma=-0.1$ are fixed here.

Subnetwork probing optimizes the mask parameter $\theta$ by minimizing the following loss function:
\begin{align*}
    &\frac{1}{|D|}\sum_{(x, y)\in D}\mathbb{E}_{U_i\sim \mathrm{Unif}[0,1]}L(f(x; \phi * z(U, \theta)), y) \\
    &+ \lambda\mathbb{E}|\theta|_0.
\end{align*}
The first term is the loss function for the model $f$ masked by $Z_i=z(U_i,\theta_i)$, and the second term corresponds to the penalty for non-zero masks to induce sparsity.
During inference, the mask $Z_i$ is binarized to $\{0, 1\}$ based on a threshold.

%% file: sections/a3_results_causal_other.tex
\input{sources/fig_results_overall_other}
\input{sources/fig_results_causal_epoch_other}
\section{Other Results of the Transition During Training}
\label{sec:causal_other}
Figure~\ref{fig:results_overall_epoch_other} shows how the model performance changes during training in \dobjppsubjpp{}, and 
Figure~\ref{fig:results_causal_epoch_other} shows the shift of the generalization performance in semantic parsing when a certain syntactic feature is removed.

%% file: sources/fig_results_overall_other.tex
\begin{figure}[t]
    \centering
    \begin{minipage}[b]{0.7\linewidth}
    \centering
    \includegraphics[width=\linewidth]{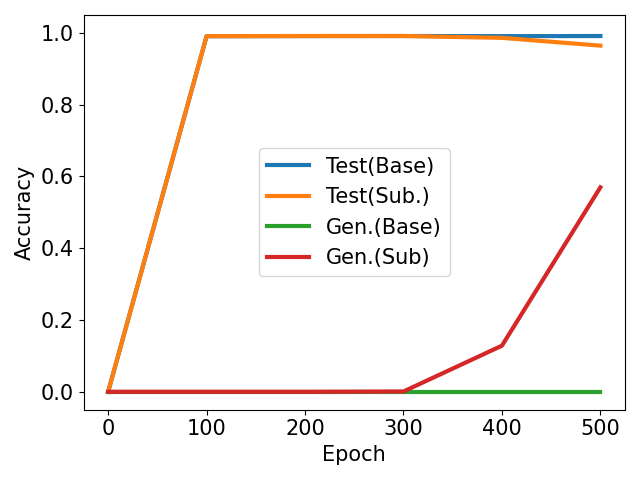}
    \subcaption{Machine translation}
    \end{minipage}
    \begin{minipage}[b]{0.7\linewidth}
    \centering
    \includegraphics[width=\linewidth]{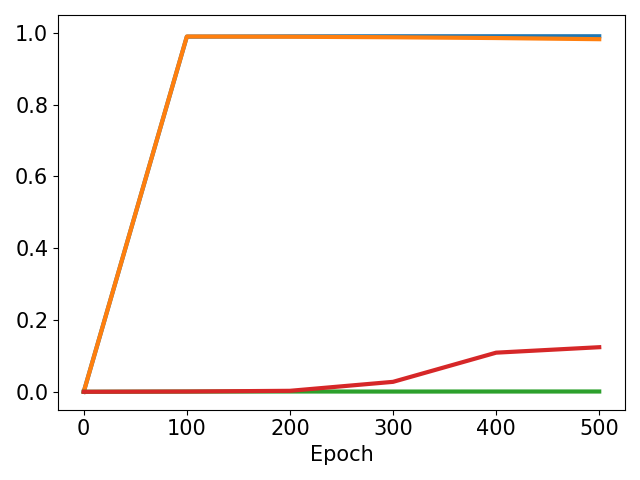}
    \subcaption{Semantic parsing}
    \end{minipage}
    \caption{Shift of average accuracy of the models in \dobjppsubjpp{} over training epochs.}
    \label{fig:results_overall_epoch_other}
\end{figure}

%% file: sources/fig_results_causal_epoch_other.tex
\begin{figure}[t]
    \centering
    \begin{minipage}[b]{0.49\linewidth}
    \centering
    \includegraphics[width=\linewidth]{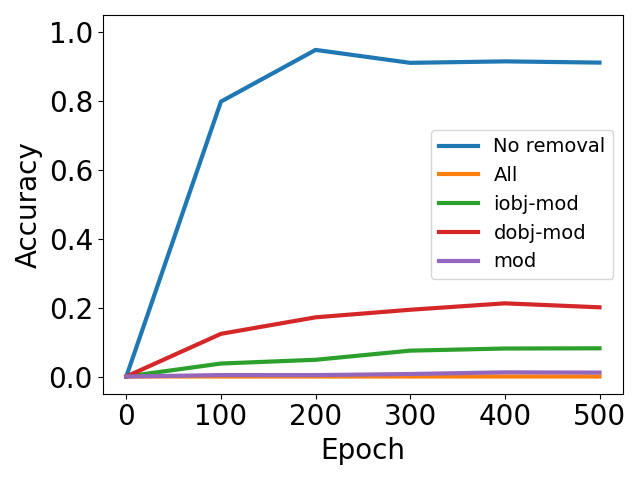}
    \subcaption{Constituency (\dobjppiobjpp{})}
    \end{minipage}
    \begin{minipage}[b]{0.49\linewidth}
    \centering
    \includegraphics[width=\linewidth]{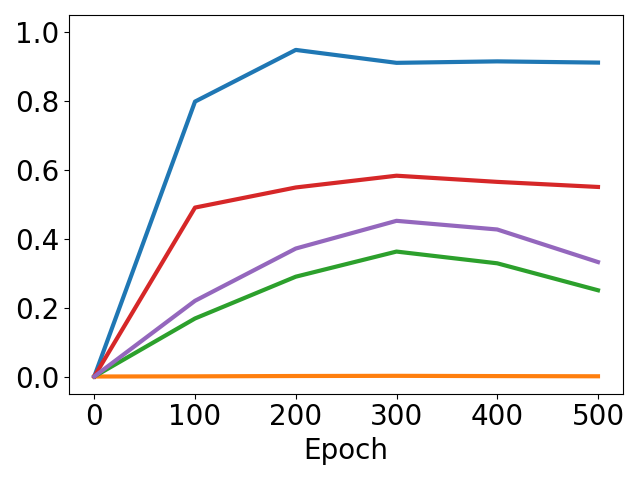}
    \subcaption{Dependency (\dobjppiobjpp{})}
    \end{minipage}
    \begin{minipage}[b]{0.49\linewidth}
    \centering
    \includegraphics[width=\linewidth]{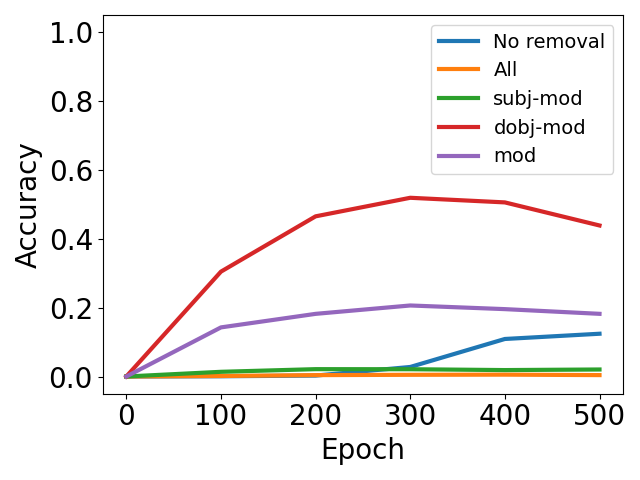}
    \subcaption{Constituency (\dobjppsubjpp{})}
    \end{minipage}
    \begin{minipage}[b]{0.49\linewidth}
    \centering
    \includegraphics[width=\linewidth]{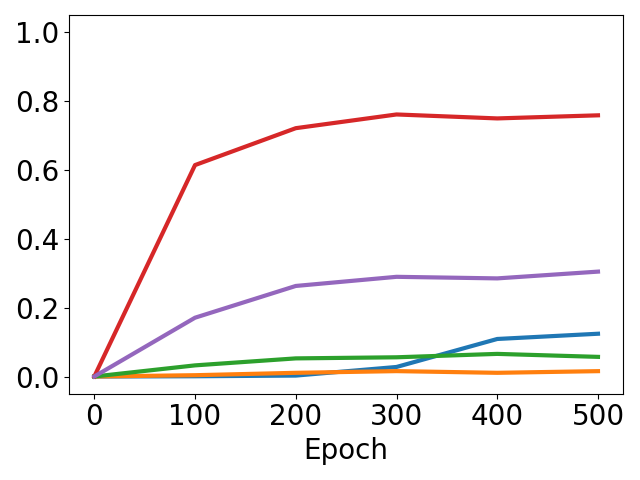}
    \subcaption{Dependency (\dobjppsubjpp{})}
    \end{minipage}
    \caption{Shift of average accuracy of the models with each concept removed in semantic parsing over training epochs.}
    \label{fig:results_causal_epoch_other}
\end{figure}

%% file: sections/a5_results_subnetwork.tex
\section{Details of Discovered Subnetworks}
\label{sec:results_subnetwork}
\input{sources/tab_results_subnetwork}
\input{sources/tab_results_subnetwork_ex}

Following \citet{cao-etal-2021-low}, we calculated the proportion of unmasked weights, and its trend was mostly the same for all tasks and patterns.
The proportion of unmasked weights in each encoder and decoder layer was around 50\%-70\%, but the proportion was larger for deeper layers in both the encoder and decoder.
Also, there were generally more unmasked weights in the decoder than in the encoder.
As for the proportion of unmasked weights in multilayer perceptron (MLP) blocks and attention blocks, most layers had more unmasked weights in MLP blocks than in attention blocks.
Table~\ref{tab:results_subnetwork} and \ref{tab:results_subnetwork_ex} show the proportions of unmasked weights in the extracted subnetworks.

%% file: sources/tab_results_subnetwork.tex
\begin{table}[t]
    \centering
    \small
    \begin{tabular}{p{0.12\linewidth}p{0.15\linewidth}p{0.15\linewidth}p{0.15\linewidth}p{0.15\linewidth}}\toprule
    Task & \multicolumn{2}{c}{MT} & \multicolumn{2}{c}{SP}\\
    Layer & \dobjppiobjpp{} Sub. & \dobjppsubjpp{} Sub. & \dobjppiobjpp{} Sub. & \dobjppsubjpp{} Sub. \\\midrule
    Enc. 0 & $55.7_{\pm 0.2}$ & $58.0_{\pm 0.3}$ & $49.8_{\pm 0.4}$ & $53.9_{\pm 0.4}$\\
    Enc. 1 & $57.9_{\pm 0.7}$ & $61.4_{\pm 0.2}$ & $51.5_{\pm 0.3}$ & $56.6_{\pm 0.4}$\\ 
    Enc. 2 & $59.6_{\pm 1.1}$ & $54.0_{\pm 0.5}$ & $63.9_{\pm 0.9}$ & $59.2_{\pm 0.2}$\\
    Dec. 0 & $59.3_{\pm 0.9}$ & $63.1_{\pm 0.6}$ & $53.5_{\pm 0.6}$ & $59.0_{\pm 0.5}$\\
    Dec. 1 & $63.8_{\pm 0.7}$ & $67.5_{\pm 0.3}$ & $58.8_{\pm 0.2}$ & $64.9_{\pm 0.2}$\\
    Dec. 2 & $68.5_{\pm 0.1}$ & $71.3_{\pm 0.2}$ & $61.6_{\pm 0.2}$ & $67.4_{\pm 0.1}$\\
    \bottomrule
    \end{tabular}
    \caption{Average proportion of unmasked weights in the two patterns and two tasks.}
    \label{tab:results_subnetwork}
\end{table}

%% file: sources/tab_results_subnetwork_ex.tex
\begin{table}[t]
    \centering
    \small
    \begin{tabular}{llll}\toprule
    Layer & Overall & Attention & MLP \\\midrule
    Enc. 0 & $55.7_{\pm 0.2}$ & $62.0_{\pm 0.3}$ & $52.5_{\pm 0.2}$\\
    Enc. 1 & $57.9_{\pm 0.7}$ & $56.8_{\pm 0.8}$ & $58.4_{\pm 1.0}$\\ 
    Enc. 2 & $59.6_{\pm 1.1}$ & $58.3_{\pm 1.3}$ & $61.0_{\pm 0.8}$\\
    Dec. 0 & $59.3_{\pm 0.9}$ & $53.9_{\pm 1.4}$ & $64.6_{\pm 0.5}$\\
    Dec. 1 & $63.8_{\pm 0.7}$ & $59.9_{\pm 1.5}$ & $67.6_{\pm 0.4}$\\
    Dec. 2 & $68.5_{\pm 0.1}$ & $64.7_{\pm 0.1}$ & $72.3_{\pm 0.2}$\\
    \bottomrule
    \end{tabular}
    \caption{Average proportion of unmasked weights in \dobjppiobjpp{} and machine translation.}
    \label{tab:results_subnetwork_ex}
\end{table}